\documentclass[runningheads]{llncs}
\usepackage{graphicx} 	
\usepackage{amsmath}
\usepackage[utf8]{inputenc} 
\usepackage{hyperref}       
\usepackage{url}            
\usepackage{booktabs}       
\usepackage{amsfonts}       
\usepackage{nicefrac}       
\usepackage{microtype}      

\usepackage{subfig}

\usepackage{amsmath}
\usepackage{graphicx}
\usepackage{xcolor}
\usepackage{etoolbox}
\usepackage[colorinlistoftodos]{todonotes}
\usepackage{makecell}

\usepackage{array}

\usepackage[utf8]{inputenc}
\usepackage{fourier} 
\usepackage{array}
\usepackage{makecell}

\usepackage{pgfplots}
\pgfplotsset{compat=1.14}
\usepackage{multirow}
\usepackage{float}

\usepackage{blindtext}

\begin{document}

\title{Trace and Detect Adversarial Attacks on CNNs using Feature Response Maps}

\author{Mohammadreza Amirian\inst{1,2} \and
Friedhelm Schwenker\inst{2} \and
Thilo Stadelmann\inst{1}}

\authorrunning{Amirian, Schwenker \& Stadelmann}
\titlerunning{Trace and Detect Adversarial Attacks on CNNs}

\institute{ZHAW Datalab \& School of Engineering, Winterthur, Switzerland \and
Institute of Neural Information Processing, Ulm University, Germany \\
\email{amir@zhaw.ch, friedhelm.schwenker@uni-ulm.de, stdm@zhaw.ch}}

\maketitle

\begin{abstract}
The existence of adversarial attacks on convolutional neural networks (CNN) questions the fitness of such models for serious applications. The attacks manipulate an input image such that misclassification is evoked while still looking normal to a human observer---they are thus not easily detectable. In a different context, backpropagated activations of CNN hidden layers---``feature responses'' to a given input---have been helpful to visualize for a human ``debugger'' what the CNN ``looks at'' while computing its output. In this work, we propose a novel detection method for adversarial examples to prevent attacks. We do so by tracking adversarial perturbations in feature responses, allowing for automatic detection using average local spatial entropy. The method does not alter the original network architecture and is fully human-interpretable. Experiments confirm the validity of our approach for state-of-the-art attacks on large-scale models trained on ImageNet.
\end{abstract}


\section{Introduction}
\label{sec:Introduction}

The success of deep neural nets for pattern recognition \cite{schmidhuber2015deep} has been a main driver behind the recent surge of interest in AI. A substantial part of this success is due to the Convolutional Neural Net (CNN) \cite{lecun1998gradient,cirecsan2011committee} and its descendants, applied to image recognition tasks. Respective methods have reached the application level in business and industry \cite{stadelmann2018beyondimagenet} and lead to a wide variety of deployed models for critical applications like automated driving \cite{bojarski2016end} or biometrics \cite{zhu2015multi}.

However, concerns regarding the reliability of deep neural networks have been raised through the discovery of so-called adversarial examples \cite{szegedy2013intriguing}. These inputs are specifically generated to ``fool'' \cite{moosavi2016deepfool} a classifier by visually appearing as some class (to humans), but being misclassified by the network with high confidence through the addition of barely visible perturbations (see Figure \ref{fig:1}). The perturbations are achieved by an optimization process on the input: the network weights are fixed, and the input pixels are optimized for the dual criterion of (a) classifying the input differently than the true class, and (b) minimizing the changes to the input. A growing body of literature confirms the impact of this discovery on practical applications of neural nets \cite{akhtar2018threat}. It raises questions on how---and in what respect different from humans---they achieve their performance, and threatens serious deployments with the possibility of tailor-made adversarial attacks.

For instance, Su et al. \cite{su2017one} report on successfully attacking neural networks by modifying a single pixel. The attack works without having access to the internal structure nor the gradients in the network under attack. Moosavi-Dezfooli et al. \cite{moosavi2017universal} furthermore show the existence of universal adversarial perturbations that can be added to any image to fool a specific model, whereas transferability of perturbations from one model to another is for example shown by Xu et al. \cite{xu2017can}. The impact of similar attacks extends beyond classification \cite{metzen2017universal}, is transferable to other modalities than images \cite{cisse2017houdini}, and also works on models distinct from neural networks \cite{papernot2016transferability}. Finally, adversarial attacks have been shown to work reliably even after perturbed images have been printed and captured again via a mobile phone camera \cite{kurakin2016adversarial}. Apparently, such research touches a weak spot.

On the other hand, there is a recent interest in the interpretability of AI agents and in particular machine learning models \cite{vellido2012making,olah2018the}. It goes hand in hand with societal developments like the new European legislation on data protection that is impacting any organization using algorithms on personal data \cite{goodman2016eu}. While neural networks are publicly perceived as ``black boxes'' with respect to how they arrive at their conclusions \cite{gunning2017explainable}, several methods have been developed recently to allow insight into the representation and decision surface of a trained model, improving interpretability. Prime candidates amongst these methods are feature visualization approaches that make the operations in hidden layers of a CNN visible \cite{zeiler2014visualizing,springenberg2014striving,olah2017feature}. They can thus serve a human engineer as a diagnostic tool in support of reasoning over success and failure of a model on the task at hand.

In this paper, we propose to use a specific form of CNN feature visualization, namely feature response maps, to not only \emph{trace} the effect of adversarial inputs on algorithmic decisions throughout the CNN;  we subsequently also use it as input to a novel automated \emph{detection} approach, based on statistical analysis of the feature responses using average of image local spatial entropy. The goal is to decide if a model is currently under attack by the given input. Our approach has the advantage over existing methods of not changing the network architecture, i.e., not affecting classification accuracy; and of being interpretable both to humans and machines, an intriguing property also for future work on the method. Experiments on the validation set of ImageNet \cite{ILSVRC15} with VGG19 networks \cite{simonyan2014very} shows the validity of our approach for detecting various state-of-the-art attacks. 

Below, Section \ref{sec:related_work} reviews related work in contrast to our approach. Section \ref{sec:background} presents the background on adversarial attacks and feature response estimation before Section \ref{sec:detection} introduces our approach in detail. Section \ref{sec:experiments} reports on experimental evaluations, and Section \ref{sec:conclusions} concludes with an outlook to future work.

\section{Related work}
\label{sec:related_work}

Work on adversarial examples for neural networks is a very active research field. Potential attacks and defenses are published at a high rate and have been surveyed recently by Akhtar and Mian \cite{akhtar2018threat}. Amongst potential defenses, directly comparable to our approach are those that focus on the sole detection of a possible attack and not on additionally recovering correct classification. 

On one hand, several detection approaches exist that exploit specific abnormal behavioral traces that adversarial examples leave while passing through a neural network: Liang et al. \cite{liang2017detecting} consider the artificial perturbations as noise in the \emph{input} and attempt to detect it by quantizing and smoothing image filters. A similar concept underlies the SqueezeNet approach by Xu et al. \cite{xu2017feature}, that compares the network's \emph{output} on the raw and filtered input, and raises a flag if detecting a large difference between both. Feinman et al. \cite{feinman2017detecting} observe the network's output confidence as estimated by dropout in the forward pass \cite{gal2016dropout}, and Lu et al's SafetyNet \cite{lu2017safetynet} looks for abnormal patterns in the ReLU activations of \emph{higher layers}. In contrast, our method performs detection based on statistics of activation patterns in the complete \emph{representation learning} part of the network as observed in feature response maps, whereas Li and Li \cite{li2016adversarial} directly observe convolutional filter statistics there.

On the other hand, a second class of detection approaches trains sophisticated classifiers for directly sorting out malformed inputs: Meng and Chen's MagNet \cite{meng2017magnet} learns the manifold of friendly images, rejects far away ones as hostile and modifies close outliers to be attracted to the manifold before feeding them back to the network under attack. Grosse et al. \cite{grosse2017statistical} enhance the output of an attacked classifier by an additional class and retrain the model to directly classify adversarial examples as such. Metzen et al. \cite{metzen2017detecting} have a similar goal but target it via an additional subnetwork. In contrast, our method uses a simple threshold-based detector and pushes all decision power to the human-interpretable feature extraction via the feature response maps.

Finally, as shown in \cite{akhtar2018threat}, different and mutually exclusive explanations for the existence of adversarial examples and the nature of neural network decision boundaries exist in the literature. Because our method enables a human investigator to trace attacks visually, it can be helpful in this debate in the future. 

\section{Background}
\label{sec:background}

We briefly present adversarial attacks and feature response estimation in general before assembling both parts into our detection approach in the next Section. 

\subsection{Adversarial attacks}
\label{sec:adversarial_attacks}

\renewcommand{\tablename}{Figure}

\begin{table}[t!]
     \begin{center}
     \begin{tabular}{  c  c  c}
     \toprule

     Original  & Difference & Adversarial \\ \midrule

      \includegraphics[width=0.14\textwidth]{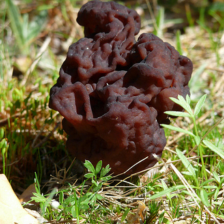} & \includegraphics[width=0.14\textwidth]{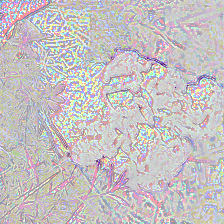} & \includegraphics[width=0.14\textwidth]{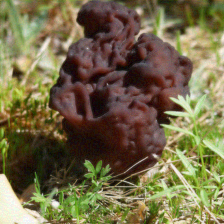}  \\ 
     Gyromitra  & Difference & Trafic light \\ \midrule
     
     \includegraphics[width=0.14\textwidth]{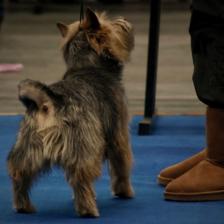} & \includegraphics[width=0.14\textwidth]{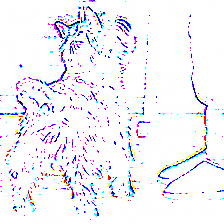} & \includegraphics[width=0.14\textwidth]{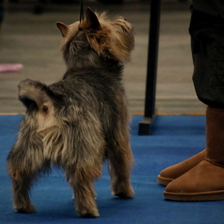}  \\ 
     Norwich terrier  & Difference & Lampshade \\ \midrule
     
      \includegraphics[width=0.14\textwidth]{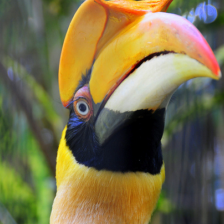} & \includegraphics[width=0.14\textwidth]{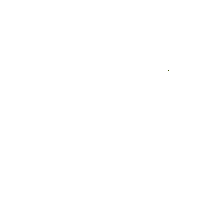} & \includegraphics[width=0.14\textwidth]{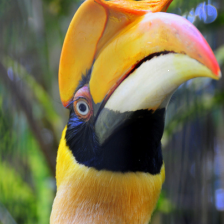}  \\ 
      Hornbill & Difference & Spotlight \\ \midrule
     
     \includegraphics[width=0.14\textwidth]{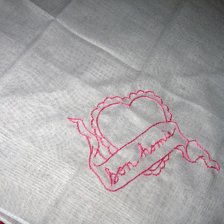} & \includegraphics[width=0.14\textwidth]{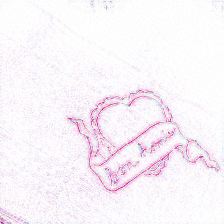} & \includegraphics[width=0.14\textwidth]{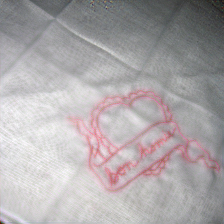} \\ 
     Handkerchief & Difference & Lampshade \\ \midrule

\includegraphics[width=0.14\textwidth]{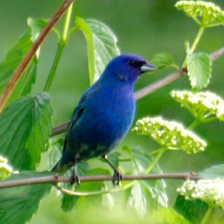} & \includegraphics[width=0.14\textwidth]{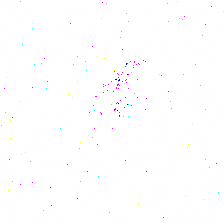} & \includegraphics[width=0.14\textwidth]{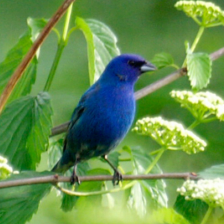}  \\ 
     Indigo bunting & Difference & Spotlight \\ \midrule     
      \end{tabular}
      \caption{Examples of different state-of-the-art adversarial attacks on a VGG19 model: original image and label (left), perturbation (middle) and mislabeled adversarial example (right). In the middle column difference of zero is encoded white and maximum difference is black because of visual enhancement.}
      \label{fig:1}
      \end{center}
      \vspace{-1.2cm}
\end{table}

The main idea of adversarial attacks is to find a small perturbation for a given image that changes the decision of the Convolutional Neural Network. Pioneering work \cite{szegedy2013intriguing} demonstrated that negligible and visually insignificant perturbations could lead to considerable deviations in the networks' output. The problem of finding a perturbation $\boldsymbol{\eta}$ for a normalized clean image $\boldsymbol{I} \in \mathbb{R}^m$, where m is the image width $\times$ height, is stated as follows \cite{szegedy2013intriguing}:
\begin{equation}
  \min_{\boldsymbol{\eta}} \parallel\boldsymbol{\eta}\parallel_2 \quad \text{s.t.} \quad \mathcal{C}(\boldsymbol{I}+\boldsymbol{\eta})\neq\ell\enspace;\quad \boldsymbol{I}+\boldsymbol{\eta} \in [0,1]^m
 \label{eq: optimization}
\end{equation}
where $\mathcal{C}(.)$ presents the classifier and $\ell\enspace$ is the ground truth label. Szegedy et al. \cite{szegedy2013intriguing} proposed to solve the optimization problem in Equation \ref{eq: optimization} for an arbitrary label $\ell^\prime$ that differs from the ground truth to find the perturbation. However, box-constrained Limmited-memory Broyden–Fletcher–Goldfarb–Shanno (L-BFGS) \cite{fletcher2013practical} is alternatively used to find perturbations satisfying Equation \ref{eq: optimization} to improve computational efficiency. Optimization based on the L-BFGS algorithm for finding adversarial attacks are computational inefficient compared with gradient-based methods. Therefore, we use a couple of gradient-based attacks, a one-pixel attack, and boundary attack to compute adversarial examples (see Figure \ref{fig:1}).

\vspace{-0.20cm}
\paragraph{\textbf{Fast Gradient Sign Method (FGSM)}} \cite{goodfellow2014explaining} is a method suggested to compute adversarial perturbations based on the gradient $\nabla_{\boldsymbol{I}}J(\boldsymbol{\theta}, \boldsymbol{I}, \ell)$ of the cost function with respect to the original image pixel values:
\begin{equation}
\boldsymbol{\eta} = \epsilon \ \text{sign} (\nabla_{\boldsymbol{I}}J(\boldsymbol{\theta}, \boldsymbol{I}, \ell))
\label{eq: FGSM}
\end{equation}
where $\boldsymbol{\theta}$ represents the network parameters and $\epsilon$ is a constant factor that constrains the max-norm $l_\infty$ of the additive perturbation $\eta$. The ground truth label is presented by $\ell$ in Equation \ref{eq: FGSM}. The $\text{sign}$ function is Equation \ref{eq: FGSM} computes the elementwise sign of the gradient of the loss function with respect to the input image. Optimizing the perturbation in Equation \ref{eq: FGSM} in a single step is called Fast Gradient Sign Method (FGSM) in the literature. This method is a white box attack, i.e. the algorithm for finding the adversarial example requires the information of weights and gradients of the network.

\vspace{-0.20cm}
\paragraph{\textbf{Gradient attack}} 
is a simple and straightforward realization of finding adversarial perturbations in the FoolBox toolbox \cite{rauber2017foolbox}. It optimizes pixel values of an ori
ginal image to minimize the ground truth label confidence in a single step.

\vspace{-0.20cm}
\paragraph{\textbf{One pixel attack}}
\cite{su2017one} is a semi-black box approach to compute adversarial examples using differential evolution \cite{storn1997differential}. The algorithm is not white box since it does not need the gradient information of the classifier; however, it is not fully black box as it needs the class probabilities. The iterative algorithm starts with randomly initialized parent perturbations. The generated offspring compete with their parent at each iteration, and the winners advance to the next step. The algorithm stops when the ground truth label probability is lower than 5\%.  

\vspace{-0.20cm}
\paragraph{\textbf{DeepFool}}
\cite{moosavi2016deepfool} is a white box iterative approach in which the closest direction to the decision boundary is computed in every step. It is equivalent to finding the corresponding path to the orthogonal projection of the data point onto the affine hyperplane which separates the binary classes. The initial method for binary classifiers can be extended to a multi-class task by considering it as multiple one-versus-all binary classifications. After finding the optimal updates toward the decision boundary, the perturbation is added to the given image. The iterations continue with estimating the optimal perturbation and apply it to the perturbed image from the last step until the network decision changes.

\vspace{-0.20cm}
\paragraph{\textbf{Boundary attack}} is a reliable black-box attack proposed by Brendel et al. in \cite{brendel2017decision}. The iterative algorithm already starts with an adversarial image and iteratively optimize the distance between this image and the original image. It
searches for an adversarial example with minimum distance from the original image.  

\subsection{Feature response estimation}
\label{sec:feature_response}

The idea of visualizing CNNs through feature responses is to find out which region of the image leads to the final decision of the network. Computing feature responses enhances the interpretability of the classifier. In this paper, we use this visualization tool to track the effect of the adversarial attacks on a CNN's decision as well as to detect perturbed examples automatically.

\begin{table}[t!]
     \begin{center}
     \begin{tabular}{  l | c  c  c c}
     \toprule
     One pixel attack \cite{su2017one}: & \includegraphics[width=0.14\textwidth]{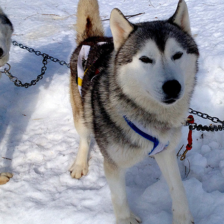} & \includegraphics[width=0.14\textwidth]{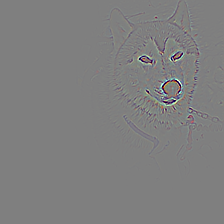} & \includegraphics[width=0.14\textwidth]{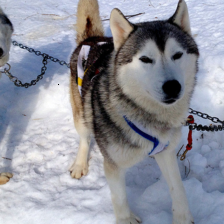} & \includegraphics[width=0.14\textwidth]{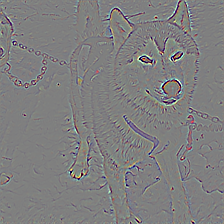} \\ 
     Predictions: & Eskimo dog & Feature response & Thimble & Feature response \\ \midrule
     FGSM \cite{goodfellow2014explaining}: & \includegraphics[width=0.14\textwidth]{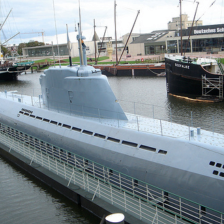} & \includegraphics[width=0.14\textwidth]{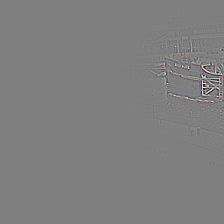} & \includegraphics[width=0.14\textwidth]{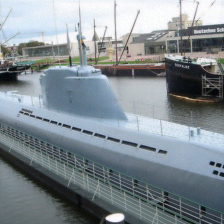} & \includegraphics[width=0.14\textwidth]{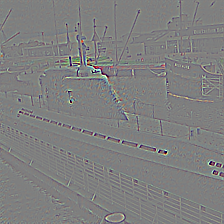}  \\ 
     Predictions: & Submarine & Feature response & Traffic light & Feature response \\ \midrule
     DeepFool \cite{moosavi2016deepfool}: & \includegraphics[width=0.14\textwidth]{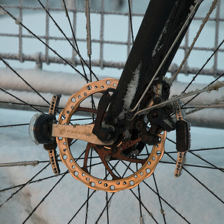} & \includegraphics[width=0.14\textwidth]{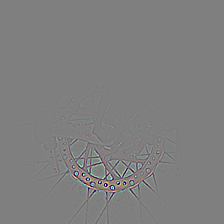} & \includegraphics[width=0.14\textwidth]{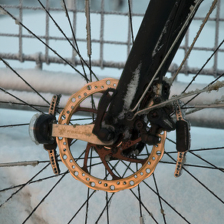} & \includegraphics[width=0.14\textwidth]{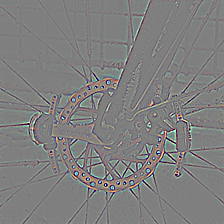} \\ 
     Predictions: & Disc brake & Feature response & Dome & Feature response \\ \toprule
      \end{tabular}
      \caption{Effect of adversarial attacks on feature responses: original image and feature response (left), perturbed versions (right).}
      \label{fig:2}
      \end{center}
      \vspace{-1.1cm}
\end{table}

Erhan et al. \cite{erhan2009visualizing} used backpropagation for visualizing feature responses of CNNs. This is implemented by evaluating an arbitrary image in the forward pass, thereby retaining the values of activated neurons at the final convolutional layer, and backpropagating these activations to the original image. The feature response has higher intensities in the regions that cause larger values of activation in the network (see Figure \ref{fig:2}). The information of max-pooling layers in the forward pass can further improve the quality of visualizations. Zeiler et al. \cite{zeiler2014visualizing} proposed to compute ``switches'', the position of maxima in all pooling regions, and then construct the feature response using transposed convolutional \cite{dumoulin2016guide} layers.

Ultimately, Springenberg et al. \cite{springenberg2014striving} proposed a combination of both methods called guided backpropagation. In this approach, the information of ``switches'' (max-pooling spatial information) is kept, and the activations are propagated backwards with the guidance of the ``switch'' information. This method leads to the best performance in network innards visualization, therefore we use guided backpropagation for computing feature response maps in this paper.

\section{Human-interpretable detection of adversarial attacks}
\label{sec:detection}

After reviewing the necessary background in the last Section, we will now present our work on tracing adversarial examples in feature response maps, which inspired a novel approach to automatic detection of adversarial perturbations in images. Using visual representations of the inner workings of neural network in this manner additionally provides a human expert guidance in developing deep convolutional networks with increased reliability and interpretability.    

\subsection{Tracing adversarial attacks in feature responses}

The research question followed in this work is to obtain insight into the reasons behind misclassification of adversarial examples. Their effect in the feature response of a CNN is for example traced in Figure \ref{fig:2}. The general phenomenon observed in all experiments is the broader feature response of adversarial examples. In contrast, Figure \ref{fig:2} demonstrates that the network looks at a smaller region of the image---is more focused---in case of not manipulated samples.

The adversarial images are visually very similar to the original ones. However, they are not correctly recognizable by deep CNNs. The original idea which triggered this study is that the focus of CNNs changes during an adversarial attack and lead to the incorrect decision. Conversely, the network makes the correct decision once it focuses on the right region of the image. Visualizing the feature response provides this and other interesting information regarding the decision making in neural networks: for instance, the image of the submarine in Figure \ref{fig:2} can be considered a good candidate for an adversarial attack since the CNN is making the decision based on an object in the background (see the feature response of the original submarine in Figure \ref{fig:2}). 

\subsection{Detecting adversarial attacks using spatial entropy}

\begin{table}[t!]
     \begin{center}
     \begin{tabular}{l | c c c c}
     \toprule
      & Original & Adversarial & Original & Adversarial \\ \midrule
      Image: & \includegraphics[width=0.14\textwidth]{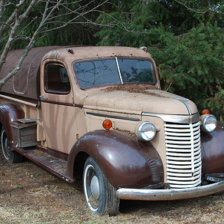} &                \includegraphics[width=0.14\textwidth]{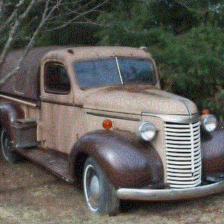} & \includegraphics[width=0.14\textwidth]       {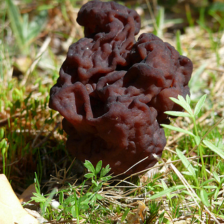} & \includegraphics[width=0.14\textwidth]{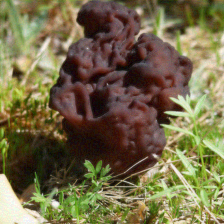} \\ \midrule
      Feature response: & \includegraphics[width=0.14\textwidth]{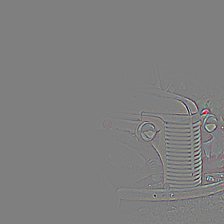} &                \includegraphics[width=0.14\textwidth]{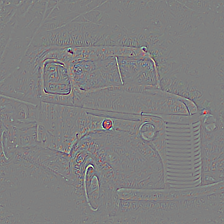} & \includegraphics[width=0.14\textwidth]       {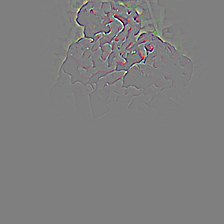} & \includegraphics[width=0.14\textwidth]{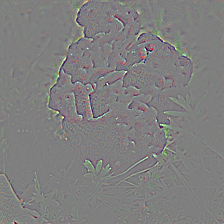} \\ \midrule
      Local spatial entropy: & \includegraphics[width=0.14\textwidth]{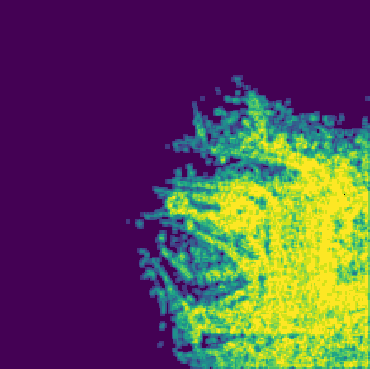} &                \includegraphics[width=0.14\textwidth]{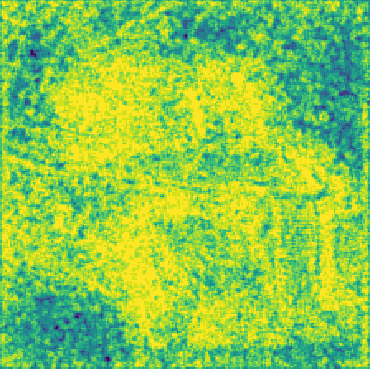} & \includegraphics[width=0.14\textwidth]       {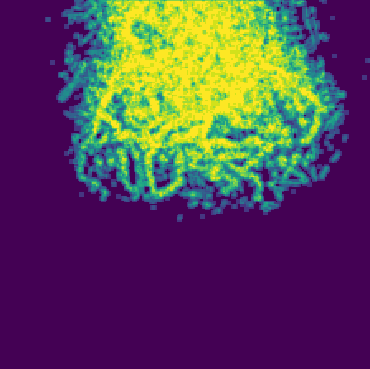} & \includegraphics[width=0.14\textwidth]{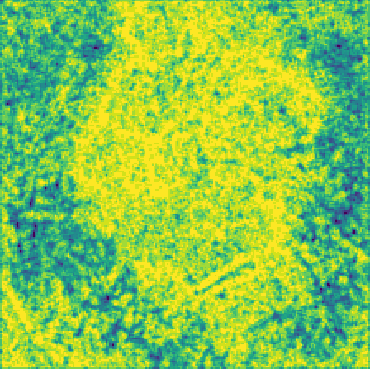} \\
      \toprule
      \end{tabular}
	  \caption{Input, feature response and local spatial entropy for clean and perturbed images, respectively.}
	  \label{fig:3}
      \end{center}
      \vspace{-1.0cm}
\end{table}

Experiments for tracing the effect of adversarial attacks on feature responses thus suggested that a CNN classifier focuses on a broader region of the input if it has been maliciously perturbed. Figure \ref{fig:2} demonstrates this connection for decision making in case of clean inputs compared with manipulated ones. The effect of adversarial manipulation is visible in the local spatial entropy of the gray-scale feature responses as well (see Figure \ref{fig:3}). The feature responses are initially converted to gray scale images, and local spatial entropies are computed based on transformed feature responses as follows \cite{chanwimaluang2003efficient}:
\begin{align}
S_k = - \sum_{i} \sum_{j} \boldsymbol{h}_k(i, j) \log_2 (h_k(i, j))
\label{eq:en}
\end{align}
where $S_k$ is the local spatial entropy of a small part (patch) of the input image and $\boldsymbol{h}_k$ represents the normalized 2D histogram value of the $k^{th}$ patch. The indices $i$ and $j$ scan through the height and width of the image patches. The patch size is $3 \times 3$ and the same as the filter size of the first layer of the used CNN (VGG19 \cite{simonyan2014very}). The local spatial entropies of corresponding feature responses are presented in Figure \ref{fig:3}, and their difference for  clean and adversarial examples suggests a likely chance to detect perturbed images based on this feature.

Accordingly, we propose to use the average local spatial entropy of an image as the final single measure to decide whether an attack has occurred or not. The average local spatial entropy $\bar{S}$ is defined as:
\vspace{-0.1cm}
\begin{align}
\bar{S} = \frac{1}{K} \sum_{k} S_k
\label{eq:en_ave}
\end{align}
\vspace{-0.1cm}
where $K$ is the number of patches on the complete feature response and $S_k$ shows the local spatial entropy as defined in Equation \ref{eq:en} and depicted in the last row of Figure \ref{fig:3}. Our detector makes the final decision by comparing the average local spatial entropy from Equation \ref{eq:en_ave} with a selected threshold, i.e., we use this feature to measure the spatial complexity of an input image (feature response).

\section{Experimental Results}
\label{sec:experiments}

To confirm the value of our final metric in Equation \ref{eq:en_ave}, we first perform experiments to visually compare the approximated distribution of the averaged local spatial entropy of feature responses in clean and perturbed images. We use the validation set of ImageNet \cite{ILSVRC15} with more than $50,000$ images from $1,000$ classes and again the VGG19 CNN \cite{simonyan2014very}. Perturbations for this experiment are computed only via the Fast Gradient Sign Attack (FGSM) method for computational reasons. Figure \ref{fig:hist_ROC}a)  shows that the clean images are separable from perturbed examples although there is some overlap between the distributions. 

\setcounter{figure}{3}
\renewcommand{\figurename}{Figure}

\begin{figure}[t]
  \centering
  \subfloat[Histogram]{{\includegraphics[width=.45\textwidth]{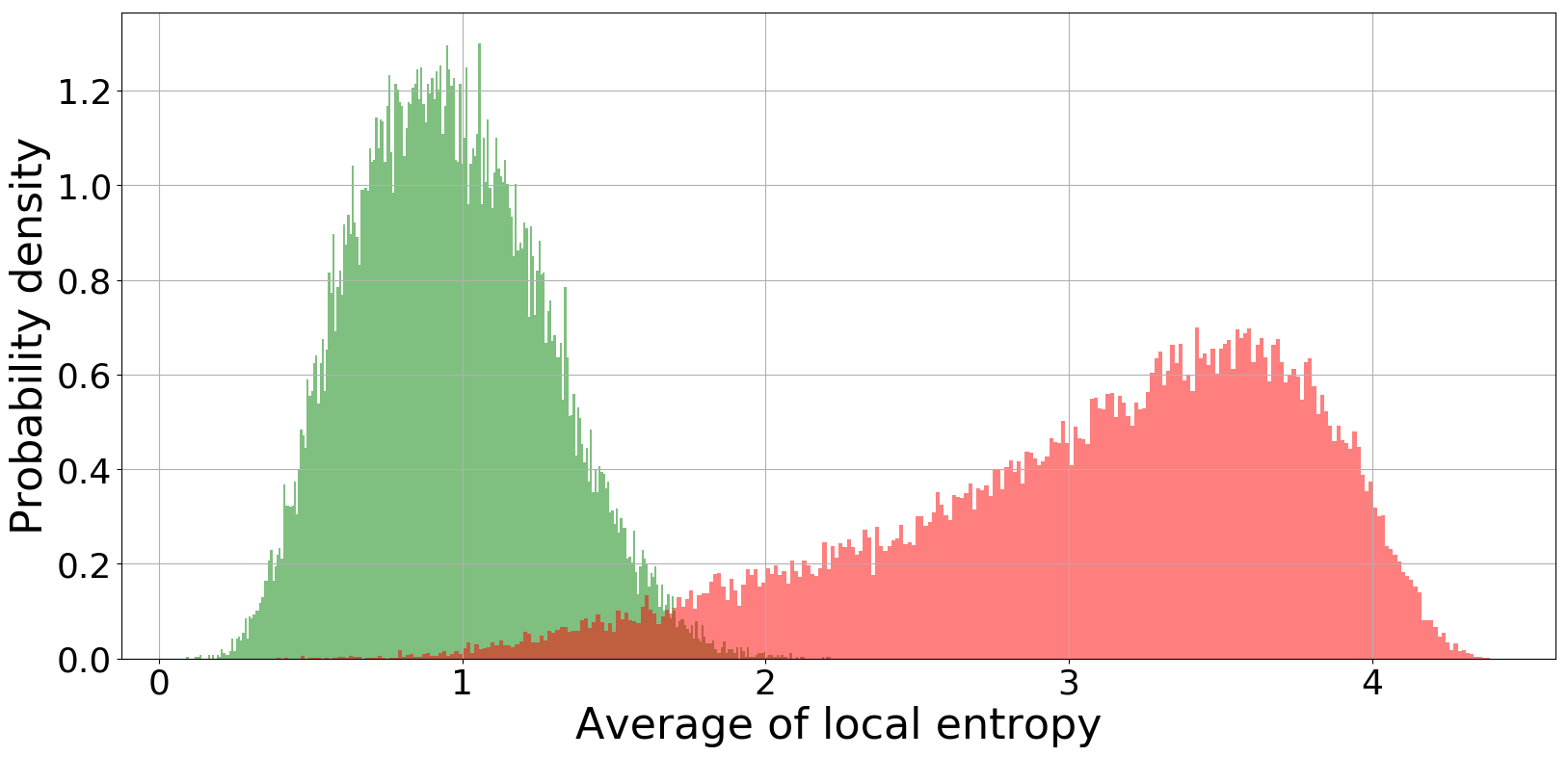} }}%
  \qquad
  \subfloat[ROC curves]{{\includegraphics[width=.45\textwidth]{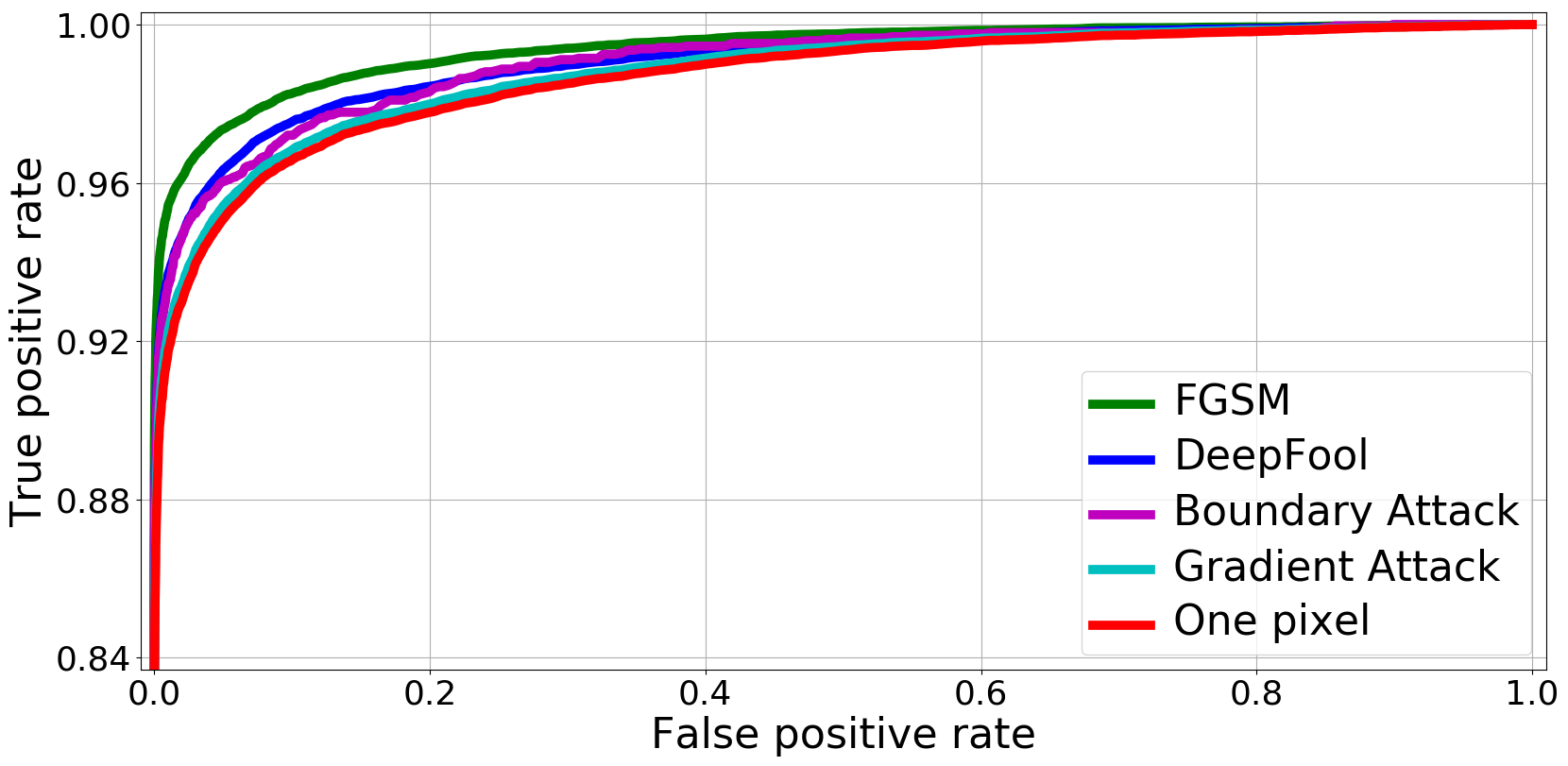} }}%
\caption{a) Distribution of average local spatial entropy in clean images (green) versus adversarial examples (red) as computed on the ImageNet validation set \cite{ILSVRC15}. b) Receiver operating characteristic (ROC) curve of the performance of our detection algorithm on different attacks.}
\vspace{+0.6cm}
  \label{fig:hist_ROC}
\end{figure}

Computing adversarial perturbations using evolutionary and iterative algorithms is demanding regarding time and computational resources. However, we would like to apply the proposed detector to a wide range of adversarial attacks. Therefore, we have drawn a number of images from the validation set of ImageNet for each attack and present the detection performance of our method in Figure \ref{fig:hist_ROC}. The selection of images is done sequentially by class and file name up to a total number of images per method that could be processed in a reasonable amount of time (see Table \ref{table:1}). We base our experiments on the FoolBox benchmarking implementation\footnote{\url{https://github.com/bethgelab/foolbox}}, running on a Pascal-based TitanX GPU.

Figure \ref{fig:hist_ROC}b presents the Receiver Operating Characteristics (ROC) of the proposed detector, and numerical evaluations are provided in Table \ref{table:1}. Our detection method performs better for gradient-based perturbations compared to the single pixel attack. Furthermore, Table \ref{table:1} suggests that the best adversarial attack detection performance is achieved for FGSM and boundary attack perturbations, where the network confidences are changed the most. This observation suggests that the proposed detector is more sensitive to attacks which are stronger in fooling the network (i.e., change the ground truth label and target class confidence more drastically). By using feature responses, we detect more than $91\%$ of the perturbed samples with a low false positive rate ($1\%$).

\setcounter{table}{0}
\renewcommand{\tablename}{Table}

\begin{table}[t!]
\begin{center}
\resizebox{\textwidth}{!}{  
\begin{tabular}{l | c | c | c | c | c  c  c}
\toprule
\multirow{2}{*}{Adversarial attack} & \#Images & \multirow{2}{*}{Success rate} &  Ground truth   &  Target class & \multicolumn{3}{|c}{False positive rate} \\
 & (run time [days]) &   &  confidence  &  confidence & 1\% & 5\% & 10\% \\
\midrule
FGSM \cite{goodfellow2014explaining}  & $50,014$ (3) & $0.925$ & $\boldsymbol{0.022}$ & $\boldsymbol{0.588}$ & $\boldsymbol{0.954}$ & $\boldsymbol{0.974}$ & $\boldsymbol{0.983}$ \\
Gradient attack \cite{rauber2017foolbox} & $50,014$ (15) & $0.499$ & $0.052$ & $0.371$ & $0.922$ & $0.954$ & $0.969$ \\ 
One pixel attack \cite{su2017one}  & $50,014$ (32) & $0.620$ & $0.037$ & $0.463$ & $0.917$ & $0.951$ & $0.966$ \\  
DeepFool \cite{moosavi2016deepfool} & $47,858$ (42) & $0.606$ & $0.041$ & $0.446$ & $0.936$ & $0.963$ & $0.976$ \\
Boundary attack \cite{brendel2017decision} & $4,013$ (17) & $\boldsymbol{0.940}$ & $0.023$ & $0.583$ & $0.934$ & $0.960$ & $0.972$ \\
\toprule
\end{tabular}
}
\end{center}
\caption{Numerical evaluation of detection performance on the three different adversarial attacks. Column two gives the amount of tested attacks and elapsed approx. run time. Success of an adversarial attack is given if a perturbation changes the prediction. Columns four and five show average confidence values of the true (ground truth) and wrong (target) class after successful attack, respectively. The last columns show detection rates for different false positive rates.}
\vspace{-0.6cm}
\label{table:1}
\end{table}

\begin{table}[t!]
\begin{center}
\resizebox{\textwidth}{!}{  
\begin{tabular}{l | c | c | c | c  c  c}
\toprule
\multirow{2}{*}{Method} &  \multirow{2}{*}{Dataset} &  \multirow{2}{*}{Network}  &  \multirow{2}{*}{Attack} & \multicolumn{3}{|c}{Performance} \\
 &   &  &  & Recall & Precision & AUC \\\midrule
Uncertainty density estimation \cite{feinman2017detecting} & SVHN \cite{krizhevsky2009learning} & LeNet \cite{lecun1989backpropagation} & FGSM & - & - & $0.890$ \\
Adaptive noise reduction \cite{liang2017detecting} & ImageNet ($4$ classes) & CaffeNet & DeepFool & $0.956$ & $0.911$  & - \\
Feature squeezing \cite{xu2017feature} & ImageNet-1000 & VGG19 & Several attacks & $0.859$ & $0.917$ & $0.942$ \\ 
Statistical analysis \cite{grosse2017statistical} & MNIST & Self-designed & FGSM ($\epsilon=0.3$) & $\boldsymbol{0.999}$ & $\boldsymbol{0.940}$ & - \\ 
Feature response (our approach) & ImageNet validation & VGG19 & Several attacks & $0.979$ & $0.920$ & $\boldsymbol{0.990}$ \\
\toprule
\end{tabular}
}
\end{center}
\caption{Performance of similar adversarial attack detection methods. The Area Under Curve (AUC) is the average value of all attacks in the third and last row.}
\label{table:2}
\end{table}

In general, it is difficult to directly compare different studies on attack detectors since they use a vast variety of neural network models, datasets, attacks and experimental setups. We present a short overview of the performances of current detection approaches in Table \ref{table:2}. Our approach is most similar to the methods of Liang et al. (\cite{liang2017detecting}) and Xu et al. (\cite{xu2017feature}). The proposed detector in this paper outperforms both based on the presented results in their work; however, we cannot guarantee identical implementations and parameterizations of the used attacks (e.g., subset of used images, learning rates for optimization of perturbations). Similarly, adaptive noise reduction in the original publication \cite{liang2017detecting} is applied to only four classes of the ImageNet dataset and defended a model based on CaffeNet, which differs from our experimental setup.

\section{Discussion and Conclusion}
\label{sec:conclusions}
The presented results demonstrate that the reality of adversarial attacks: improving the robustness of CNNs is necessary. However, we conducted further preliminary experiments on binary (cat versus dog \cite{parkhi2012cats}) and ternary (among three classes of cars \cite{krause20133d}) classification tasks as proxies for the kind of few-class classifications settings frequently arising in practice. They suggest that it is more challenging to find adversarial examples in such a setting without plenty of ``other classes'' to pick from for misclassification. Figure \ref{fig:6} illustrates these results.

\setcounter{table}{5}
\renewcommand{\tablename}{Figure}

\begin{table}[t!]
     \begin{center}
     \begin{tabular}{ c  c  c  c}
     \toprule
     Original & Adversarial & Original & Adversarial \\ \midrule
     \includegraphics[width=0.14\textwidth]{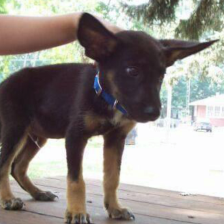} &             \includegraphics[width=0.14\textwidth]{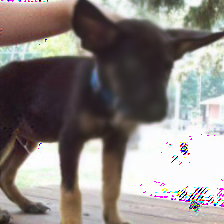} & \includegraphics[width=0.14\textwidth]{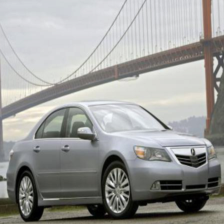} & \includegraphics[width=0.14\textwidth]{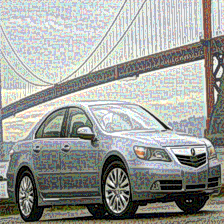}  \\ 
     \toprule
      \end{tabular}
      \caption{Successful adversarial examples created by DeepFool \cite{moosavi2016deepfool} for binary and ternary classification tasks are only possible with notable visible perturbations.}
      \label{fig:6}
      \end{center}
      \vspace{-1.0cm}
\end{table}

In this paper, we have presented an approach to detect adversarial attacks based on human-interpretable feature response maps. We traced the effect of adversarial perturbations on the visual focus of the network in original images, which inspired a simple yet robust approach for automatic detection. This proposed method is based on thresholding the averaged local spatial entropy of the feature response maps and detects at least $91\%$ of state-of-the-art adversarial attacks with a low false positive rate on the validation set of ImageNet. However, the results are not directly comparable with methods in the literature because of the diversity in the experimental setups and implementations of attacks. 

Our results verify that feature response are informative to detect specific cases of failure in deep CNNs. The proposed detector applies to increase the interpretability of neural network decisions, which is an increasingly important topic towards robust and reliable AI. Future work, therefore, will concentrate on developing reliable and interpretable image classification methods for practical use cases based on our preliminary results for binary and ternary classification.

\paragraph{\textbf{Acknowledgements}}
We are grateful for the support by Innosuisse grant 26025.1 PFES-ES ``QualitAI''.
\vspace{-0.1cm}
\bibliography{annpr2018-2}

\begin{thebibliography}{10}
\providecommand{\url}[1]{\texttt{#1}}
\providecommand{\urlprefix}{URL }
\providecommand{\doi}[1]{https://doi.org/#1}

\bibitem{akhtar2018threat}
Akhtar, N., Mian, A.: Threat of adversarial attacks on deep learning in
  computer vision: A survey. arXiv preprint arXiv:1801.00553  (2018)

\bibitem{bojarski2016end}
Bojarski, M., Del~Testa, D., Dworakowski, D., Firner, B., Flepp, B., Goyal, P.,
  Jackel, L.D., Monfort, M., Muller, U., Zhang, J., et~al.: End to end learning
  for self-driving cars. arXiv preprint arXiv:1604.07316  (2016)

\bibitem{brendel2017decision}
Brendel, W., Rauber, J., Bethge, M.: Decision-based adversarial attacks:
  Reliable attacks against black-box machine learning models. arXiv preprint
  arXiv:1712.04248  (2017)

\bibitem{chanwimaluang2003efficient}
Chanwimaluang, T., Fan, G.: An efficient blood vessel detection algorithm for
  retinal images using local entropy thresholding. International Symposium on
  Circuits and Systems (ISCAS)  \textbf{5} (2003)

\bibitem{cirecsan2011committee}
Cire{\c{s}}an, D., Meier, U., Masci, J., Schmidhuber, J.: A committee of neural
  networks for traffic sign classification. In: IJCNN. pp. 1918--1921. IEEE
  (2011)

\bibitem{cisse2017houdini}
Cisse, M., Adi, Y., Neverova, N., Keshet, J.: Houdini: Fooling deep structured
  prediction models. arXiv preprint arXiv:1707.05373  (2017)

\bibitem{dumoulin2016guide}
Dumoulin, V., Visin, F.: A guide to convolution arithmetic for deep learning.
  arXiv preprint arXiv:1603.07285  (2016)

\bibitem{erhan2009visualizing}
Erhan, D., Bengio, Y., Courville, A., Vincent, P.: Visualizing higher-layer
  features of a deep network. University of Montreal  \textbf{1341}(3), ~1
  (2009)

\bibitem{feinman2017detecting}
Feinman, R., Curtin, R.R., Shintre, S., Gardner, A.B.: Detecting adversarial
  samples from artifacts. arXiv preprint arXiv:1703.00410  (2017)

\bibitem{fletcher2013practical}
Fletcher, R.: Practical methods of optimization. John Wiley \& Sons (2013)

\bibitem{gal2016dropout}
Gal, Y., Ghahramani, Z.: Dropout as a bayesian approximation: Representing
  model uncertainty in deep learning. In: ICML (2016)

\bibitem{goodfellow2014explaining}
Goodfellow, I.J., Shlens, J., Szegedy, C.: Explaining and harnessing
  adversarial examples. ICLR  (2015)

\bibitem{goodman2016eu}
Goodman, B., Flaxman, S.: Eu regulations on algorithmic decision-making and a
  “right to explanation”. In: ICML workshop on human interpretability in
  machine learning (WHI) (2016)

\bibitem{grosse2017statistical}
Grosse, K., Manoharan, P., Papernot, N., Backes, M., McDaniel, P.: On the
  (statistical) detection of adversarial examples. arXiv preprint
  arXiv:1702.06280  (2017)

\bibitem{gunning2017explainable}
Gunning, D.: Explainable artificial intelligence ({XAI}). Defense Advanced
  Research Projects Agency (DARPA)  (2017)

\bibitem{krause20133d}
Krause, J., Stark, M., Deng, J., Fei-Fei, L.: {3D} object representations for
  fine-grained categorization. In: ICCV Workshops (2013)

\bibitem{krizhevsky2009learning}
Krizhevsky, A., Hinton, G.: Learning multiple layers of features from tiny
  images  (2009)

\bibitem{kurakin2016adversarial}
Kurakin, A., Goodfellow, I., Bengio, S.: Adversarial examples in the physical
  world. ICRL Workshop track  (2016)

\bibitem{lecun1989backpropagation}
LeCun, Y., Boser, B., Denker, J.S., Henderson, D., Howard, R.E., Hubbard, W.,
  Jackel, L.D.: Backpropagation applied to handwritten zip code recognition.
  Neural computation  \textbf{1}(4),  541--551 (1989)

\bibitem{lecun1998gradient}
LeCun, Y., Bottou, L., Bengio, Y., Haffner, P.: Gradient-based learning applied
  to document recognition. Proceedings of the IEEE  \textbf{86}(11),
  2278--2324 (1998)

\bibitem{li2016adversarial}
Li, X., Li, F.: Adversarial examples detection in deep networks with
  convolutional filter statistics. arXiv preprint arXiv:1612.07767  (2016)

\bibitem{liang2017detecting}
Liang, B., Li, H., Su, M., Li, X., Shi, W., Wang, X.: Detecting adversarial
  examples in deep networks with adaptive noise reduction. arXiv preprint
  arXiv:1705.08378  (2017)

\bibitem{lu2017safetynet}
Lu, J., Issaranon, T., Forsyth, D.: Safetynet: Detecting and rejecting
  adversarial examples robustly. arXiv preprint arXiv:1704.00103  (2017)

\bibitem{meng2017magnet}
Meng, D., Chen, H.: Magnet: a two-pronged defense against adversarial examples.
  In: ACM SIGSAC Conference on Computer and Communications Security (2017)

\bibitem{metzen2017detecting}
Metzen, J.H., Genewein, T., Fischer, V., Bischoff, B.: On detecting adversarial
  perturbations. ICLR  (2017)

\bibitem{metzen2017universal}
Metzen, J.H., Kumar, M.C., Brox, T., Fischer, V.: Universal adversarial
  perturbations against semantic image segmentation. arXiv preprint
  arXiv:1704.05712  (2017)

\bibitem{moosavi2017universal}
Moosavi-Dezfooli, S.M., Fawzi, A., Fawzi, O., Frossard, P.: Universal
  adversarial perturbations. arXiv preprint arXiv:1610.08401  (2017)

\bibitem{moosavi2016deepfool}
Moosavi~Dezfooli, S.M., Fawzi, A., Frossard, P.: Deepfool: a simple and
  accurate method to fool deep neural networks. In: CVPR (2016)

\bibitem{olah2017feature}
Olah, C., Mordvintsev, A., Schubert, L.: Feature visualization. Distill
  (2017). \doi{10.23915/distill.00007}

\bibitem{olah2018the}
Olah, C., Satyanarayan, A., Johnson, I., Carter, S., Schubert, L., Ye, K.,
  Mordvintsev, A.: The building blocks of interpretability. Distill  (2018).
  \doi{10.23915/distill.00010}

\bibitem{papernot2016transferability}
Papernot, N., McDaniel, P., Goodfellow, I.: Transferability in machine
  learning: from phenomena to black-box attacks using adversarial samples.
  arXiv preprint arXiv:1605.07277  (2016)

\bibitem{parkhi2012cats}
Parkhi, O.M., Vedaldi, A., Zisserman, A., Jawahar, C.: Cats and dogs. In: CVPR
  (2012)

\bibitem{rauber2017foolbox}
Rauber, J., Brendel, W., Bethge, M.: Foolbox v0.8.0: A python toolbox to
  benchmark the robustness of machine learning models. arXiv preprint
  arXiv:1707.04131  (2017)

\bibitem{ILSVRC15}
Russakovsky, O., Deng, J., Su, H., Krause, J., Satheesh, S., Ma, S., Huang, Z.,
  Karpathy, A., Khosla, A., Bernstein, M., Berg, A.C., Fei-Fei, L.: Imagenet
  large scale visual recognition challenge. International Journal of Computer
  Vision (IJCV)  \textbf{115}(3),  211--252 (2015).
  \doi{10.1007/s11263-015-0816-y}

\bibitem{schmidhuber2015deep}
Schmidhuber, J.: Deep learning in neural networks: An overview. Neural networks
   \textbf{61},  85--117 (2015)

\bibitem{simonyan2014very}
Simonyan, K., Zisserman, A.: Very deep convolutional networks for large-scale
  image recognition. ICLR  (2015)

\bibitem{springenberg2014striving}
Springenberg, J.T., Dosovitskiy, A., Brox, T., Riedmiller, M.: Striving for
  simplicity: The all convolutional net. arXiv preprint arXiv:1412.6806  (2014)

\bibitem{stadelmann2018beyondimagenet}
Stadelmann, T., Tolkachev, V., Sick, B., Stampfli, J., D\"urr, O.: Beyond
  imagenet - deep learning in industrial practice. In: Braschler, M.,
  Stadelmann, T., Stockinger, K. (eds.) Applied Data Science - Lessons Learned
  for the Data-Driven Business. Springer (2018), to appear

\bibitem{storn1997differential}
Storn, R., Price, K.: Differential evolution--a simple and efficient heuristic
  for global optimization over continuous spaces. Journal of global
  optimization  \textbf{11}(4),  341--359 (1997)

\bibitem{su2017one}
Su, J., Vargas, D.V., Kouichi, S.: One pixel attack for fooling deep neural
  networks. arXiv preprint arXiv:1710.08864  (2017)

\bibitem{szegedy2013intriguing}
Szegedy, C., Zaremba, W., Sutskever, I., Bruna, J., Erhan, D., Goodfellow, I.,
  Fergus, R.: Intriguing properties of neural networks. ICLR  (2014)

\bibitem{vellido2012making}
Vellido, A., Mart{\'\i}n-Guerrero, J.D., Lisboa, P.J.: Making machine learning
  models interpretable. In: ESANN. vol.~12, pp. 163--172 (2012)

\bibitem{xu2017feature}
Xu, W., Evans, D., Qi, Y.: Feature squeezing: Detecting adversarial examples in
  deep neural networks  (2018)

\bibitem{xu2017can}
Xu, X., Chen, X., Liu, C., Rohrbach, A., Darell, T., Song, D.: Can you fool
  {AI} with adversarial examples on a visual turing test? arXiv preprint
  arXiv:1709.08693  (2017)

\bibitem{zeiler2014visualizing}
Zeiler, M.D., Fergus, R.: Visualizing and understanding convolutional networks.
  In: ECCV (2014)

\bibitem{zhu2015multi}
Zhu, J., Liao, S., Yi, D., Lei, Z., Li, S.Z.: Multi-label cnn based pedestrian
  attribute learning for soft biometrics. In: International Conference on
  Biometrics (ICB). IEEE (2015)

\end{thebibliography}
\bibliographystyle{splncs04}

\end{document}